# Urdu & Hindi Poetry Generation using Neural Networks

Urdu & Hindi Poetry Generation

A system to overcome writer's block by suggesting machine-generated prompts for aspiring poets.

**Shakeeb A. M. Mukhtar**

Computer Science & Engineering, Vishwakarma Institute of Technology, Pune, Maharashtra, India, shakeeb.ahmad19@vit.edu

**Pushkar S. Joglekar**

Computer Science & Engineering, Vishwakarma Institute of Technology, Pune, Maharashtra, India, pushkar.joglekar@vit.edu

One of the major problems writers and poets face is the *writer's block*. It is a condition in which an author loses the ability to produce new work or experiences a creative slowdown. The problem is more difficult in the context of poetry than prose, as in the latter case authors need not be very concise while expressing their ideas, also the various aspects such as rhyme, poetic meters are not relevant for prose. One of the most effective ways to overcome this writing block for poets can be, to have a prompt system, which would help their imagination and open their minds for new ideas. A prompt system can possibly generate one liner, two liner or full ghazals. The purpose of this work is to give an ode to the Urdu/Hindi poets, and helping them start their next line of poetry, a couplet or a complete ghazal considering various factors like rhymes, refrain, and meters. The result will help aspiring poets to get new ideas and help them overcome writer's block by auto-generating pieces of poetry using Deep Learning techniques. A concern with creative works like this, especially in the literary context, is to ensure that the output is not plagiarized. This work also addresses the concern and makes sure that the resulting odes are not exact match with input data using parameters like temperature and manual plagiarism check against input corpus.

To the best of our knowledge, although the automatic text generation problem has been studied quite extensively in the literature, the specific problem of Urdu/Hindi poetry generation has not been explored much. Apart from developing system to auto-generate Urdu/Hindi poetry, another key contribution of our work is to create a cleaned and preprocessed corpus of Urdu/Hindi poetry (derived from authentic resources) and making it freely available for researchers in the area.

## 1 INTRODUCTION

The writer's block a poet faces can be quite frustrating. The constraints of Urdu poetry such as meter, rhymes, refrain, underlying mood etc. make the problem even more challenging. Nevertheless, the heart of the problem is coming up with the new idea underlying the piece of poetry. While writing ghazals, choosing a random rhyme and refrain is sometimes used as an outlet from the creative blockage, but often it might not result into a fresh idea. We found that, a random yet unique verse inferred via our neural network-based approach turns out to be effective in generating a novel idea. Once poet lands on an interesting idea, the other aspects such as meter, rhymes, refrain etc., are not very difficult to take care of. More often, an experienced poet can find suitable synonymous word (sometimes semantically corelated words) which when used appropriately can transform the generated poetry into well-formed ghazal which follows rules of rhymes, meter, refrain etc.

In the literature, the problem of automatic text generation has been explored quite extensively. It has several applications ranging from simple email replies and word suggestions to writing in the styles of Shakespeare and simulating DNA sequences. The former is relevant for the current project. The aim for this study is to overcome writer's block by suggesting machine-generated prompts for aspiring poets in Urdu/Hindi. While a lot of work for generating poetry using Machine Learning has been done in other languages, Indic languages like Urdu/Hindi are still far behind despite being culturally rich. Very few works have been done for the specific problem, mostly due to lack of a good dataset.

A quality corpus is often overlooked but is of crucial importance in text-generation projects. We prudently selected, and later fine-tuned our resources for the raw poetry data under the supervision of domain experts. The primary source is Rekhta [21] – an Indian literary web portal owned by Rekhta Foundation; a nonprofit NGO dedicated to the promotion of Urdu literature in South Asia. The portal provides as-is content from original sources and serves it in multiple scripts such as Devanagari, Roman and, primarily, Perso-Arabic.

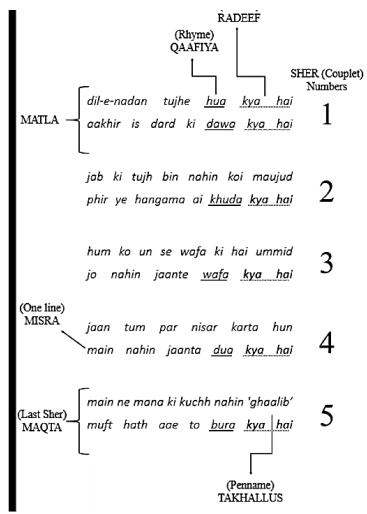

Figure 1: Formal features of a Ghazal illustrated with reference to a ghazal by the Urdu poet Ghalib (d. 1869)

The whole system of this work has been built around *ghazal*. Explaining its structure, Christopher Shackle describes [20] that a ghazal consists of a series of rhymed verses (*she'r*), each symmetrically divided into two half-verses (*misra'*). The above ghazal shows five verses, usually reckoned the minimum number needed to make up a complete ghazal, with about a dozen being the normal maximum. The verses of the ghazal are quite often described in English as 'couplets', but the implied analogy with English poetry is somewhat misleading. The formal structure (*tarah, zamin*) of a ghazal is defined by its metre or meter (*behr*) and its rhyme (*qaafiya*), which are both maintained consistently throughout the poem. Thus, each half-verse is written in the same metrical pattern of alternating short (S) and long (L) syllables. There are about half a dozen Persian-derived ghazal meters in standard use. The meter used here is called *khafeef*, where each half-verse has the metrical pattern LS LL SL SL L L, denoted in the traditional mnemonic system of meaningless syllables (*afaa'iil*) as *fa'ilaatun mafaa'ilun fi'lun.*

Each full verse is end-stopped, with the ending marked by a highly defined rhyme, in this case the polysyllabic *-a kya hai,* which would be divided by the traditional rhetoric into the qaafiya or 'rhyme proper' –*a* and the radif or 'end-rhyme' or 'refrain' *kya hai*. The prominence of the rhyme is underlined by its repetition at the end of both halves of the opening verse (matla'). The ending of the poem is marked by the inclusion in the final verse (maqta') of the poet's pen-name (takhallus), a Persian word adopted as signature.

## 2 RELATED WORK

Before choosing a suitable model for our purpose, we studied similar previous works and compared them. The techniques, models and approaches used were thoroughly examined for their efficiency considering their domain of work, i.e., prose or poetry. Later, we adapted some of the techniques in the context of our problem.

*Charcter-level RNN:* Generating text character by character is found to be quite effective [2]. The results are astonishing as there are no language models used to train on. The model in a way learns itself the language of the training dataset, without explicit use of language models. This makes the approach particularly effective in case of non-English based corpus, as the training dataset available are predominantly for English and not for Indic languages. Naturally, since language models are not used, this approach sometimes may produce result which may not satisfy certain constraints imposed by grammar rules at "global" level (sentence level). As we noted earlier, it does not reduce effectivity of the approach for our problem as generating a novel idea is the goal and further enhancement to bring it into meter, rhyme is more of a routine task which can be easily accomplished by experienced poets.

*Seq2Seq:* A 2018 approach to the problem is by Neueste Beiträge [1] on a readily available Hindi corpus. After cleaning up the corpus, it is transliterated into roman (English text). The transliteration module used is open-source and the output quality is not examined as the author, apparently, is not familiar with the Devanagari script or Hindi language. The transliterated outcome is then modified to feed in for Neural Network training.

After creating a sequence-to-sequence model, the final output is generated, citing the best ones. The generated results are not rated, verified, or improved by any means.

*Transformer-based generative models:* According to Rakesh Verma and Avisha Das [3] from University of Houstan, automated long content generation is a difficult task maintaining coherence becomes more challenging with an increase in the length. Deep neural architectures such as Recursive Neural Networks (RNNs), Long Short Term Memory (LSTM) networks are widely used for content generation owing to their ability to learn dependencies across the textual context. LSTMs have been used for generating stories, Shakespearean Sonnets, and non-English poetry, and for reading comprehension-based tasks.

Comparing different techniques themselves, [3] notice that massive textual content has enabled rapid advances in natural language modeling. The use of pre-trained deep neural language models has significantly improved natural language understanding tasks. However, the extent to which these systems can be applied to content generation is unclear.

While a few informal studies have claimed that these models can generate 'high quality' readable content, there is no prior study on analyzing the generated content from these models based on sampling and fine-tuning hyperparameters. They conduct an in-depth comparison of several language models for open-ended story generation from given prompts. Using a diverse set of automated metrics, we compare the performance of transformer-based generative models – OpenAI's GPT2 (pre-trained and fine-tuned) and Google's pretrained Transformer-XL and XLNet to human-written textual references. Studying inter-metric correlation along with metric ranking reveals interesting insights – the high correlation between the readability scores and word usage in the text.

A study of the statistical significance and empirical evaluations between the scores (human and machine-generated) at higher sampling hyperparameter combinations reveal that the top pre-trained and fine-tuned models generated samples condition well on the prompt with an increased occurrence of unique and difficult words.

The GPT2-medium model fine-tuned on the 1024 Byte-pair Encoding (BPE) tokenized version of the dataset along with pre-trained Transformer-XL models generated samples close to human written content on three metrics: prompt-based overlap, coherence, and variation in sentence length. A study of overall model stability and performance shows that fine-tuned GPT2 language models have the least deviation in metric scores from human performance.

A similar approach was used by Kushal Chauhan [17] who experimented with RNNs and later open AI's GPT-3 trained on Hindi corpus and fine-tuned on verses of Hindi poetry.

*RNN & LSTM:* The system proposed by Dipti Pawade [6] aims to generate a new story based on a series of inputted stories. For new story generation, they have considered two possibilities with respect to nature of inputted stories. Firstly, they have considered the stories with different storyline and characters. Secondly, they have worked with different volumes of the same stories where the storyline is in context with each other, and characters are also similar.

Results generated by the system are analyzed based on parameters like grammar correctness, linkage of events, interest level and uniqueness.

*RNN and Skip Grams:* Chandra et al. [9] in their work have proposed a Natural Language Processing and Deep Learning based algorithmic framework which is demonstrated by generating the context for products on e-commerce websites. Their procedure involves primarily 5 steps carried out in an unsupervised manner which are implemented using extraction and abstraction.

The steps include –
   a) collecting keywords and description for a particular item for data aggregation
   b) avoiding redundancy by removing duplicate and item-specific information by creating a blacklist dictionary
   c) filtering those sentences and paragraphs that satisfy the user's intent

    d) using Recurrent Neural Network (RNN) along with Long Short-Term Memory (LSTM) for sentence generation and Skip Grams with Negative Sampling for extraction

    e) implementing TextRank so that the output is ordered contextually to provide a coherent read.

*Statistical models & ConceptNet:* In a similar context, Thomaidou et al. [10] have proposed a method to produce promotional text snippets of advertisement content by considering the landing page of the website as their input.

To articulate a piece of a promotional snippet, they have followed the following steps:

    a) information extraction
    b) sentiment analysis
    c) natural language generation

Parag Jain et al. [11], discussed a model which takes sequences of short narrations and generates a story out of it. They have used phrased based Statistical Machine Translation to figure out target language phrase-maps and then merge all the phrase-maps to get the target language text. For learning and sequence generation sequence-to-sequence recurrent neural network architecture [12] is implemented.

Richard et al. [13] described an expert writing prompt application. Here three random words are selected from WordNet. Concepts of selected words are examined by using ConceptNet and then concept correlation is carried out to set the theme. Characters are set using plot generation. Here iteratively new ConceptNet search is carried out to build the full-fledged plot. Lastly, story evaluation is done and if it is not as per the goal, the plot is modified iteratively.

## 2.1 Comparison of the studies

The approach used in [1] cannot be considered very effective or thorough as the author is not familiar with the language, thus restraining him to assist the results during training by supervision. For example:

- Correct the corpus before training
- Pre-processing the corpus correctly
- Fixing Transliteration mistakes which leads to unnecessarily incorrect grammar
- Verifying the results

The study by Avisha et. el. [3] summarizes various strategies. The parameters considered by [6] are not necessarily best measures, but a general starting point for evaluation of such results.

The story generation methods using sequential models as in [11] or full-fledged prompt-based generation like [10] and [13] are mostly based on hand-written if-then rules. Although rule-based approaches are proved to be effective in machine learning because they rely heavily on human supervision, it can also endlessly complicate the tasks when the rule-list continue to expand.

Transformer based models used by [3] and [17] gives promising results but the fine-tuning the models using low-quality datasets can result into substandard outcomes, compared to classics and sophisticated pieces of poetry which follows meters and conventional poetic devices.

The automatic language-learning approaches are maybe less effective, but most of the time work more efficiently based on runtime, scalability, learning curve and inference.

Although a significant amount of work in English has been done regarding the natural language text (poetry) generation, there is almost no work available in Urdu or Hindi.

No language models are present specifically for Indic languages. While a lot of work for generating poetry using Machine Learning has been done in other languages, Indic languages like Urdu/Hindi are still far behind despite being culturally rich. No datasets are readily available in the case of Indic language poetry generation. We had to scrape data from websites to collect Urdu/Hindi poetry and then pre-process them before using it for training our model. We have made both raw and cleaned dataset created during the work freely available for future works [19].

## 3 SYSTEM DESIGN AND CONSTRAINTS

We have used a character-level RNN with LSTM, which has been proved to be a promising approach for text generation in Indic languages. The available datasets are mostly for English, in our case we had to build the dataset from scratch by scraping data from various websites. In case of a recurring neural network, that too based on a character-level language model, we need a quality dataset, and an LSTM based multi-traversal model to train our data.

Note that, the desired output should not be exactly the text used for the training, so there must be some means to tweak and control the randomness exhibited by the output text. It is observed that, word-level models often produce grammatically consistent output but there is no evident way to control randomness observed in the output and often resulting output shows significant match with the input text used. Whereas, in case of character-level models the output might not always be grammatically correct, but one can control the randomness (e.g., through temperature parameter we used in our work) and it results in increased stochasticity. Moreover, it is observed that with sufficient epochs the output obtained using character-level model can be transformed to grammatically correct output with small changes.

**CONSTRAINTS**

To keep the work concise, we identified following constraints:

1. The focus is to give a prompt to the aspiring poets and assist them with writer's block. Thus, the output will have a unique idea inferred from the trained data models with Refrain and Rhymes.
2. The output *may* not be –
   - Fully consistent with grammar rules
   - In poetic meter

Despite the built language models' abilities, which can possibly get these parameters right, it is not guaranteed. Although the approaches to counter these limitations are also discussed in this study, it is not of the prime focus. Our basic roadmap is as follows:

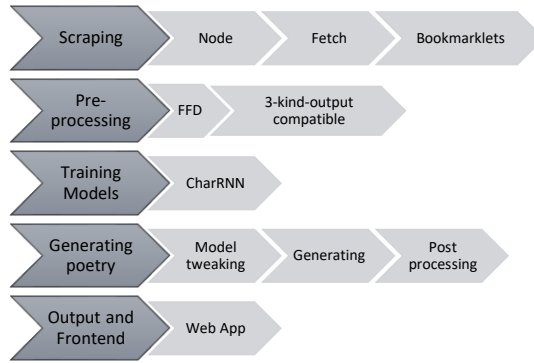

Figure 2: Implementation Roadmap

After scraping the data from multiple credible sources, it will be pre-processed to be fed in the 128-cell 3-layer LSTM network. The model is tweaked for UTF-8 character encodings, trained and then used to generate poetry using command-line instructions or the designed frontend. The output can then be further assessed and scrutinized using post-processing modules like meter-check and plagiarism-check.

### 3.1 Corpus Collection using Data Scraping

The quality of corpora is an important feature affecting their usability in NLP applications. Currently no decent dataset is available, and a cleaned high-quality dataset collected under the supervision of domain experts will be extremely useful for future development.

To collect poetry corpus and make a sufficiently large dataset, we wrote several scripts and built custom tools for data scraping from rekhta.org and several other standard sites like UrduWeb's Urdu Mehfil and Bazm-e-Urdu Library [22].

For grabbing the links to later fetch and scrape, we created custom JavaScript bookmarklets. The links were then fed to the Node.js script scraping ghazals of various poets available on Rekhta.org, collecting them in text files. Firstly, we tried puppeteer, a headless chrome which runs at the backend. The 'puppeteer' approach proved to be very slow compared to Node.js fetch and manual parsing with cheerio. Making the script more adaptive, we modified it to save contents in all three scripts: English, Hindi, and Urdu.

All the scraped raw data has been made freely available for future researchers [19].

### 3.2 Data Pre-processing

The scraped data is formatted and aligned to make it compatible with the neural network.

As the end-result should offer 3 modes to the user, the input dataset should fulfill 3-output compatibility to fit the model. The Neural Networks will be trained on these datasets separately and the resulting language models will ultimately be used for the output.

These 3 kinds of output would be *Misra'*, *She'r* and *Ghazal*.

*1. One liner (called misra'):* Will be obtained by splitting the input on newline character

*2. Two liner (called she'r):* Will be obtained by splitting the input on every second line (odd-even logic)

*3. Full ghazal:* Will be obtained by splitting the input on special mark entered during scraping

### 3.3 Training a Char-RNN on the data

Character Level Recurring Neural Networks, as opposed to word level RNNs, take one character at a time (Urdu/Hindi alphabets in our case) and create vectors on them.

Various RNN libraries are available, including Char-RNNs, for various tasks. The text-generation by char-RNN has been found out to be unreasonably effective. [2]

The default model we used takes in up to 40 characters of input, converts every character to a 100-D character embedding vector, and feeds those into a 128-cell LSTM recurrent layer. The three stacked layers use previous output as an input and then are then fed into an Attention layer to consider the most significant temporal features and average them. All the numeric parameters above can be configured in a custom model.

We decided to train the corpus splitting it into 80-20 or 70-30 for training & testing respectively, whichever works best, using Keras library in Python. Number of epochs will be decided based on training and validation error. If the validation error starts increasing, that might be an indication of overfitting. Initially we used 20 epochs, then increased it up to 100 and noticed the changes. To boost the training process, we used Google Collaboratory with a Tesla K80 GPU.

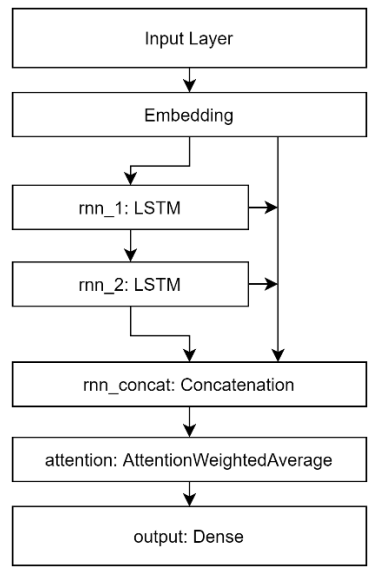

Figure 2: RNN model for the system

### 3.4 Generating Poetry – one/two liners or Ghazals

The trained models can then be used to generate texts (poetry, in this case) of variant lengths.

One liner (called *misra'*), by splitting on newline character, will generate more unique piece of poetry, and will be not-so-much-similar-to-the-original.

We speculated that it would have less semantical consistency as a demerit. Also, it will be independent to any ghazal, thus rhyme (called *qaafiya*), refrain (called *radiif*) and in some cases, even meter will not be fixed as we cannot determine rhymes based on only one line. Considering the prime focus of the study, i.e., to overcome writer's block, even such outputs will be helpful, as meter and refrain are less-creative aspects compared to the idea.

Two liners (called *she'r*) will be similar, but it will respect the refrain and rhyme (and meter, to some extent). It will be more semantically related. Still, randomness will be high, which is good for the prompts generated for aspiring poets, just to give new ideas, but bad for highly meaningful poetry.

Full Ghazal will generate a long consistent piece of poetry, with rhymes and refrain. A demerit (if considered one) will be that the result would be more like the original text.

### 3.5 Output and Front-end

The generated piece of poetry (from python) was exposed to the user/frontend. A web-app prototype was made, as it can be converted into apps for any platform. For that, we used:

Eel - A little Python library for making simple Electron-like offline HTML/JS GUI apps, with full access to Python capabilities and libraries.

Ionic – A framework for quick UI prototyping with material design approach

Miscellaneous: Cordova for Android app, Electron for Desktop app

### 3.6 Extra

To assist the generated prompts, we added a module to calculate rhythm and meter of generated poetry and rating it accordingly. This further helps the aspiring poets to understand and write better poetry.

Calculating meter (*behr*) is a difficult task and out of the scope of our work, but a similar work [15] has been done in Urdu and we used its API service in our front-end.

To check for the plagiarism and avoid using exact same prompts from the input corpus, we implemented an anti-plagiarism module. If matches are found, they are listed for the user to decide whether to select the machine-generated prompt or not.

## 4 RESULTS

### 4.1 Phase One

To test the model initially, we used a dataset comprising of poetry of 41 poets in Devanagari script. After training for 20 epochs on a 5.5 MB file containing 61620 lines, 1 *misra* out of 9 was acceptable either directly or with slight changes. even 2-3 sometimes.

By this time, the scraper we wrote has been evolved, and we decided to open-source it for the world [18][19] along with bookmarklets. It was a better version which did not require puppeteer and could be customized for both prose and poetry.

*Sample Outputs*

Table 1: Sample output verses in Phase 1

| Verse | Meter | Accept/Reject | | |
|---|---|---|---|---|
| | | Grammar | Idea | Final |
| हमारे जिस्म के बाज़ार कहने लगते हैं | ✓ | ✓ | ✓ | ✓ |

| Verse | Meter | Accept/Reject | | |
|---|---|---|---|---|
| | | Grammar | Idea | Final |
| *Hamaare jism ke bazaar kehne lagte hain* | | | | |
| ऐ 'ज़फ़र' उस बुत-ए-नाशाद की बात पर<br>*Ae zafar us but-e-nashaad ki baat par* | ✓ | ✓ | ✓ | ✓ |
| किस क़दर अपने दिल की है ये कि तारीफ़<br>*Kis qadar apne dil ki hai ye ki ta'reef* | ✗ | ✓ | ✓ | ✗ |
| इस दिल में है वो मरहम-ए-साहिल कहाँ<br>*Is dil men hai wo marham-e-saahil kahaN* | ✓ | ✗ | ✓ | ✓ |

With slight modifications, the results which were not in the meter and/or semantically incorrect were easy to fix.

*Example 1:* जैसे कुछ इस आज़ार से बातें होंगे

Original:    *jaise kuchh is aazaar se baten honge*

Changed:   *kuchh in aazaar se baten hongi*

[There will be talks using such lame excuses]

*Example 2:* कि मुझ को न कहता है कि मैं कुछ नहीं कहता

Original:    *ki mujh ko na kehta hai ki main kuchh nahin kehta*

Changed:   *vo mujh ko ye kehta hai ke main kuchh nahin kehta*

[They say unto me that why do not I say anything]

### 4.2 Phase Two

Scraper was updated, and more data was added into the corpus. Meanwhile, the Devanagari data of usable quality could not be found anymore, and we faced problems scraping available Urdu data on few other sites.

Going with the final Devanagari data, we trained it for 100 epochs. We noticed that more training could lead to perfection, which in turn generates output very similar to the input. We introduced a parameter "temperature" in the training model to generate the output with desired randomness. From 0 being super random to 1 being perfect, we studied the outputs and with human supervision, we found 0.8 as most efficient temperature.

Significant improvements were noticed, as 6-9 out of 10 generated prompts were acceptable.

A user interface was designed using Ionic framework and Vanilla JavaScript to use different models, generate, view, and analyze the outputs in an efficient way. Eel framework was used as an interface to the python backend.

We also implemented two extra modules to assist in improving the generated output:

*Transliteration:* From Hindi to Urdu & vice versa, using Patiala university tool *Sangam* [14].

Meter: Checks the meter of the generated output according to traditional Perso-Arabic metric system devised by Khalil Ahmad. This works only on Urdu text, but using the transliteration module discussed above as an interceptor, it works for Devanagari too. The API used is from aruuz.com by Syed Zeeshan [15].

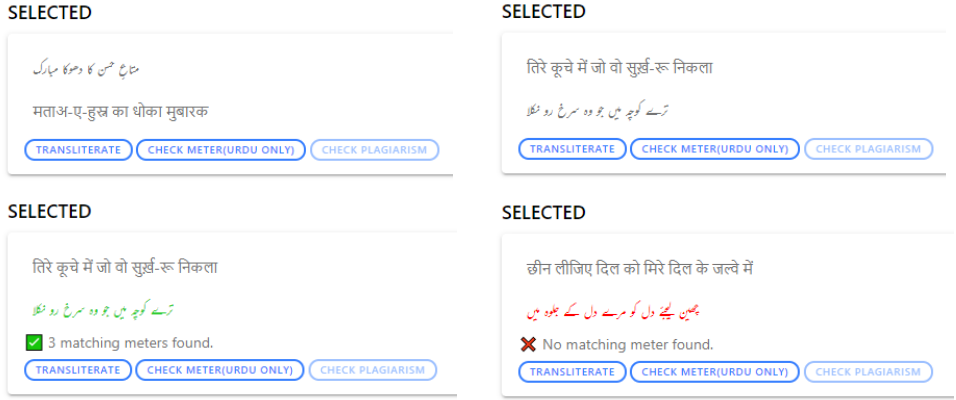

Figure 3: Transliteration – Urdu-Hindi-Urdu and Poetic Meter Check

*Sample Outputs*

Table 2: Sample output verses in Phase 2

| Verse | Meter | Accept/Reject | | |
|---|---|---|---|---|
| | | Grammar | Idea | Final |
| न हो जाए बे-ज़बाँ में कोई<br>*Na ho jae be-zubaaN meN koi* | ✓ | ✓ | ✓ | ✓ |
| तिरे कूचे में जो वो सुर्ख़-रू निकला<br>*Tere kooche meN jo wo surkhru nikla* | ✓ | ✓ | ✓ | ✓ |
| तो क्या ये बात करते हैं<br>*To kya ye baat karte haiN* | ✓ | ✓ | ✗ | ✗ |
| बे-ख़ुदी को जो मैं ने किस का यार किया<br>*Be-khudi ki jo maiN ne kis ka yaar kiya* | ✓ | ✗ | ✓ | ✓ |
| कि वो जब तक है तो क्या है जो मुझ को<br>*Ki wo jab tak hai to kya hai jo mujh ko* | ✓ | ✗ | ✓ | ✗ |

Examples of 'accepted' verses with slight modifications:

*Example 1:* छीन लीजिए दिल को मिरे दिल के जल्वे में

Original: *chheen lijiye dil ko <u>mere</u> dil ke jalwe <u>men</u>*

Changed: *chheen lijiye dil ko apne dil ke jalwe se*

 Or  *qaid kijiye dil ko apne dil ke zindaan men*

[Enchain my heart in the prison of your heart]

*Example 1:* बंदा-ए-अब्र-ए-तन्हाई का जवाब हुआ

Original: *banda-e-<u>abr-e</u>-tanhai ka jawaab hua*

Changed: *banda-e-ishq ki tanhai ka jawaab hua*

[It became the response to the loneliness of man of love]

Looking at few results, the domain experts concluded that some of the outputs can come out as plain silly, but ultimately, they too can be useful. For instance, generated output "रू-ए-यार को तो कमर करे" *(roo-e-yaar ko to kamar kare)* could easily be picked up and with a slight push of words, an average poet can come up with something like: *Behtar hai roo-e-yaar ko ab tu qamar kahe*. This tremendously increases the usability and efficiency of our work.

### 4.3 Phase Three

*Plagiarism Module*

The plagiarism check was implemented successfully and included in the web-app. The app now:
1. Lets you select any text on the screen, grabs it and then check the selected text for the plagiarism.
2. Number of matching lines exactly as in the training dataset.
3. Returns all the lines for manual review.

This improved the user satisfaction and efficiency further.

We found that a total match of a generated output is almost impossible. The partial searches using selected words can assist the user though.

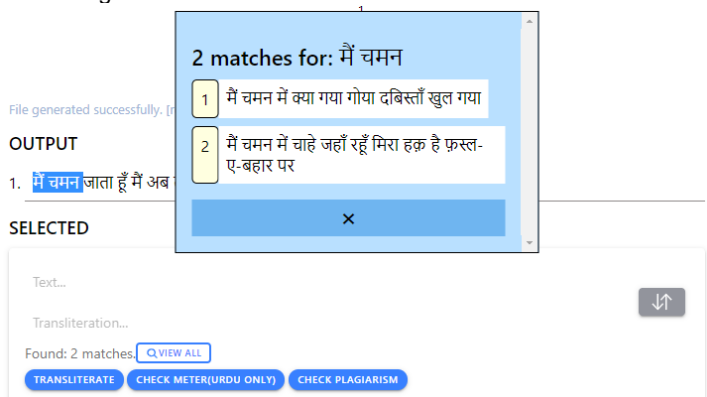

Figure 4: Plagiarism Check in action

We started scraping more data before training the available Urdu dataset. In total, we grabbed the following:
- 153749 lines before clean-up
- 6993 Ghazals
- 116376 misre
- Removed 8928 lines had English text
- 123386 lines (or 61693 couplets/sher) after clean-up
- 7MB file

For the command window, i.e., without the UI, we also had it possible to get involved in how the output unfolds, step by step. Interactive mode will suggest you the top N options for the next word and allows you to pick one.

For example, after initial work दिल, we get a prompt to select from top, say, 5 words according to the trained model. It will say:

Choose next word:

1. में
2. ने
3. से
4. को
5. अपना

On choosing a word from the list, say, 3, we get next five words suggested to us and so on.

We also added options like prefix and temperature to get desired output starting from a string and adjust its 'randomness', respectively. In LSTMs, temperature increases the sensitivity to low probability candidates. The candidate, or sample in our case is a letter. Using a higher temperature makes the RNN more "easily excited" by samples, resulting in more diversity.

Figure 5: Updated UI with more options

### Noticeable advantages of Urdu script

We noticed that the compound words (*taraakiib*) were even more interesting in Urdu, because the long vowel representing -e- or -ए-in Hindi, which is called 'zer-e-izaafat' in Persian and Urdu poetry, is absent. The absence of *zer-e-izaafat* leaves the results a bit ambiguous, (see short-vowels in Arabic and Urdu [16]) and the user can imagine any *tarkiib* of his choice.

### Hindi Vs Urdu

The results were significantly improved in Urdu script compared to Hindi, both because the actual script was to be Urdu, and because of extra data we scraped for it.

We speculate that if the training poetry data were for a language more leaning-towards-Sanskrit Hindustani, the Devanagari script would have given better results.

### Acceptance Ratio

About 9-14 results out of 20 were acceptable based on the predefined parameters.

*Test Cases*

- Sampled via Random Sampling, wrote a script to randomly select any N lines
- Noticed acceptable outputs per 20 sampled input
- 4 passes to take algebraic Mean

*Validation Approach*

- Semi-automated
    - Automated: Meter, Plagiarism check
    - Manual: Idea, Grammar
- Accept/Reject based on rhythm, musicality, meter, syntax (spellings), semantics (grammar)

*Sample Outputs*

Table 3: Sample output verses in Phase 3

| Verse | Meter | Accept/Reject | | |
| --- | --- | --- | --- | --- |
| | | Grammar | Idea | Final |
| फिर उस की बातों के हो गए हैं<br>*Phir us ki baton ke hog ae hain* | ✓ | ✓ | ✓ | ✓ |
| चले आते हैं उस के सादा-ए-दिल से सब कुछ हो<br>*Chale aate hain us ke sada-e-dil se sab kuchh ho* | ✓ | ✗ | ✓ | ✓ |
| वो ज़मीं से है मंज़िल-ए-मह-ओ-ताब<br>*Vo zamiin se hai manzil-e-mah-o-taab* | ✓ | ✓ | ✓ | ✓ |
| शब-ए-फ़िराक़ में जो आए तो हम ने<br>*Shab-e-firaaq men jo aae to ham ne* | ✓ | ✓ | ✓ | ✓ |
| फ़ित्ना-कार देख रहे हैं वो बस<br>*Fitna-kaar dekh rahe hain vo bas* | ✓ | ✗ | ✓ | ✗ |

## 5 CONCLUSION & FUTURE WORK

Text generation is one of the machine learning's application that is commonly used for businesses and academia alike. Writing story prompts and generating text in a specific style were achieved successfully in the past. The Indic languages though did not make it in the race of technical linguistic advances.

Our project demonstrates a way to overcome writers' block by suggesting machine generated prompts to the users. The results were found to be quite impressive, considering the specific use-case of taking someone out of writer's block. Most of the outputs being generated were already in poetic meter without any explicit respective provision. With aid of additional modules like meter checks, transliteration and plagiarism, the efficiency improved further.

The test results and statistical analysis used in this report did not consider this added benefit. If done so with related parameters, other relatively substandard results are guaranteed to go from acceptable to impressive.

We enlist the features of our system:
1. Generate poetry prompts for the aspiring poets with customized temperature, output length, prefix and two scripts – Urdu & Hindi.
2. Distraction-free UI for post-processing and presentation of individual output from several generated prompts.
3. A module to transliterate the text from Urdu to Hindi and vice versa.
4. A module to calculate poetic meters depicting if the output fits a meter or not and shows the matching meters if it does.
5. A module to check plagiarism in generated output against original dataset, and a list of matching verses.

Our approach, even though simple, turns out to be quite effective in generating poetry which is acceptable by domain experts. Also, the first author of this paper himself is a poet (see his blogs[*] for his works) with a formal knowledge of Urdu poetic devices and teaches aspiring poets via workshops, so the supervision of the work in this aspect was assuring all along.

We believe, our work is just a starting point and there are several interesting directions which can be explored further.
1. One can provide Grammar check using conditional models.
2. In case of Character level RNNs, metrics like edit distance can be used to correct the words or to show suggestions.
3. If one wants to generate longer outputs, the character-level RNNs directly might not be useful, as the output might not be easily transformable to grammatically correct one. So, a plausibility of somehow combining "local" (char level RNN) and "global" (based on grammar, language models) models can be explored.
4. Also, the problem of introducing a parameter to control the randomness in the output of language model-based approaches is quite interesting. One possibility is: the sequence in which production rules of the grammar are applied can be governed with suitable parameter.
5. When one would like to generate poetry with couple of lines, word embeddings or sentence embeddings can possibly be used to ensure that the different sentences generated are semantically correlated to some extent.
6. Using an adaptive model to supervise it in run time, so that the "selections" a user makes influences the next generated output.
7. **Making feedback of the network more sensitive to specific parameters such as meter score, or the extent to which the output is grammatically correct, etc.**

---

*\* https://www.shakeeb.in and https://ur.shakeeb.in*